\documentclass[11pt,a4paper]{article}
\usepackage[hyperref]{emnlp2018}
\usepackage{times}
\usepackage{latexsym}
\usepackage{multirow}
\usepackage{graphicx}
\usepackage{paralist} 
\usepackage{amsfonts}
\usepackage{amsmath}
\usepackage{amsthm}

\usepackage{url}
\usepackage{soul}
\usepackage{tabu}
\usepackage[normalem]{ulem}
\setuldepth{Berlin}

\aclfinalcopy 

\newcolumntype{L}[1]{>{\raggedright\let\newline\\\arraybackslash\hspace{0pt}}m{#1}}
\newcolumntype{C}[1]{>{\centering\let\newline\\\arraybackslash\hspace{0pt}}m{#1}}
\newcolumntype{R}[1]{>{\raggedleft\let\newline\\\arraybackslash\hspace{0pt}}m{#1}}

\usepackage[scaled=.90]{helvet} 

\title{Topic Memory Networks for Short Text Classification}

\author{Jichuan Zeng$^{1}$\thanks{~~This work was mainly conducted when Jichuan Zeng was an intern in Tencent AI Lab.}, Jing Li$^{2}$\thanks{~~Jing Li is the corresponding author.}, Yan Song$^2$, Cuiyun Gao$^1$, Michael R. Lyu$^1$, Irwin King$^1$ \\
  $^1$ Department of Computer Science and Engineering \\The Chinese University of Hong Kong, HKSAR, China \\
  $^2$ Tencent AI Lab, Shenzhen, China \\
  $^1$ {\tt \{jczeng,cygao,lyu,king\}@cse.cuhk.edu.hk}\\
  $^2$ {\tt \{ameliajli,clksong\}@tencent.com}
  }

\date{}

\begin{document}
\maketitle
\begin{abstract}
Many classification models work poorly on short texts due to data sparsity.
To address this issue, we propose \textsl{topic memory networks} for short text classification with a novel topic memory mechanism to encode
latent topic representations indicative of class labels.
Different from most prior work that focuses on extending features with external knowledge or pre-trained topics, our model jointly explores topic inference and text classification with memory networks in an end-to-end manner.
Experimental results on four benchmark datasets show that our model outperforms state-of-the-art models on short text classification, meanwhile generates coherent topics.
\end{abstract}

\section{Introduction}\label{sec:intro}

Short texts have become an important form for individuals to voice opinions and share information on online platforms.
A large body of daily-generated contents, such as tweets, web search snippets, news feeds, and forum messages, have far outpaced the reading and understanding capacity of individuals.
As a consequence, there is a pressing need for automatic language understanding techniques for processing and analyzing such texts~\cite{DBLP:conf/naacl/ZhangLSZ18}.
Among those techniques, text classification is a critical and fundamental one proven to be useful in various downstream applications, such as text summarization \cite{DBLP:conf/emnlp/HuCZ15}, recommendation \cite{DBLP:journals/tist/ZhangWXZ12}, and sentiment analysis \cite{DBLP:conf/emnlp/ChenSBY17}.

Although many classification models like 
support vector machines (SVMs)~\cite{DBLP:conf/acl/WangM12} and 
neural networks
~\cite{DBLP:conf/emnlp/Kim14,DBLP:journals/corr/XiaoC16,joulin2017bag} have demonstrated their success in processing formal and well-edited texts, such as news articles~\cite{DBLP:conf/nips/ZhangZL15}, their performance is inevitably compromised when directly applied to short and informal online texts.
This inferior performance is attributed to the severe data sparsity
nature of short texts, which results in the
limited features available for classifiers~\cite{DBLP:conf/www/PhanNH08}.
%
To alleviate the data sparsity problem, some approaches exploit knowledge from external resources like Wikipedia \citep{DBLP:conf/cikm/JinLZYY11} and knowledge bases (\citealp{DBLP:conf/cikm/LuciaF14,DBLP:conf/ijcai/WangWZY17}).
These approaches, however, rely on a large volume of high-quality external data, which may be unavailable to some specific domains or languages (\citealp{DBLP:conf/sigir/LiWZSM16}).
\begin{table}\small
\centering
\begin{tabular}{|p{7.2cm}|}
\hline
\underline{\textbf{Training instances}}\\
R$_1$: $[$\textsf{SuperBowl}$]$ I'll do anything to see the \textbf{Steelers} win. \\
R$_2$: $[$\textsf{New.Music.Live}$]$ Please give \textbf{wristbands}, she have major \textbf{Bieber} Fever.\\
\hline
\hline
\underline{\textbf{Test instance}}\\
S: $[$\textsf{New.Music.Live}$]$ I will do anything for \textbf{wristbands}, gonna tweet till I win.\\
\hline
\end{tabular}
\caption{Tweet examples for 
classification. R$_i$ denotes the i-th training instance; S denotes a test instance. $[$\textsf{class}$]$ is the ground-truth label. \textbf{Bold} words are indicative of an instance's class label.}\label{tab:intro-example}
\vskip -0.5em
\end{table}

To illustrate the difficulties in classifying short texts, we take the tweet classification in Table~\ref{tab:intro-example} as an example. In the test instance S, only given the $11$ words it contains, it is difficult to understand why its label is \textsf{New.Music.Live}. Without richer context, classifiers are likely to classify S into the same category as the training instance R$_1$, which happens to share many words with S, in spite of the different categories they belong to,\footnote{R$_1$ is about \textsf{SuperBowl}, the annual championship game of the National Football League. R$_2$ and S are both about \textsf{New.Music.Live}, the flagship live music show.} rather than R$_2$, which only shares the word ``\textit{wristbands}'' with S. 
Under this circumstance, how might we enrich the context of these short texts? 
If looking at R$_2$, we can observe that the semantic meaning of ``\textit{wristbands}'' can be extended from its co-occurrence with ``\textit{Bieber}'', 
which is highly indicative of \textsf{New.Music.Live}.\footnote{Justine Bieber was on New.Music.Live in 2011. There was a business activity for this event that gave free wristbands to fans if they supported Bieber on Twitter.}
Such relation can further help in recognizing the word ``\textit{wristbands}'' to be important when classifying the test instance S.

%

Motivated by the above-mentioned observations, we present a novel neural framework, named as \textsl{topic memory networks} (TMN), for short text classification that does not rely on external knowledge.
Our model can identify the indicative words for classification, e.g., ``\textit{wristbands}'' in S, via jointly exploiting the document-level word co-occurrence patterns, e.g., ``\textit{wristbands}'' and ``\textit{Bieber}'' in R$_2$. 
To be more specific, built upon the success of neural topic models~\cite{srivastava2017autoencoding,DBLP:conf/icml/MiaoGB17}, our model is capable of
discovering \textsl{latent topics}\footnote{ Latent topics are the distributional clusters of 
words that frequently co-occur in some of the instances instead of widely appearing throughout the corpus \cite{DBLP:journals/jmlr/BleiNJ03}.}, which can capture the co-occurrence of words in document level.
To employ the \textsl{latent topics} for short text classification, we propose a novel \textsl{topic memory mechanism}, which is inspired by memory networks~\cite{DBLP:journals/corr/WestonCB14,DBLP:journals/corr/GravesWD14}, that allows the model to put attention upon the indicative latent topics useful to classification. With such corpus-level latent topic representations, each short text instance is enriched, which thus helps alleviate the data sparsity issues.

In prior research, though the effects of topic models for short text classification have been explored~\cite{DBLP:conf/www/PhanNH08,DBLP:conf/aaai/RenZZJ16a}, existing methods tend to use pre-trained topics as features. To the best of our knowledge, our model is the first to encode latent topic representations via memory networks for short text classification, which allows joint inference of latent topics.
To evaluate our model, we experiment and compare it with existing methods on four benchmark datasets.
Experimental results indicate that our model outperforms state-of-the-art counterparts on short text classification.
The quantitative and qualitative analysis illustrate the capability of our model in generating topic representations that are meaningful and indicative of different categories.

\begin{figure}[t]
	\centering
	\includegraphics[width=0.48 \textwidth]{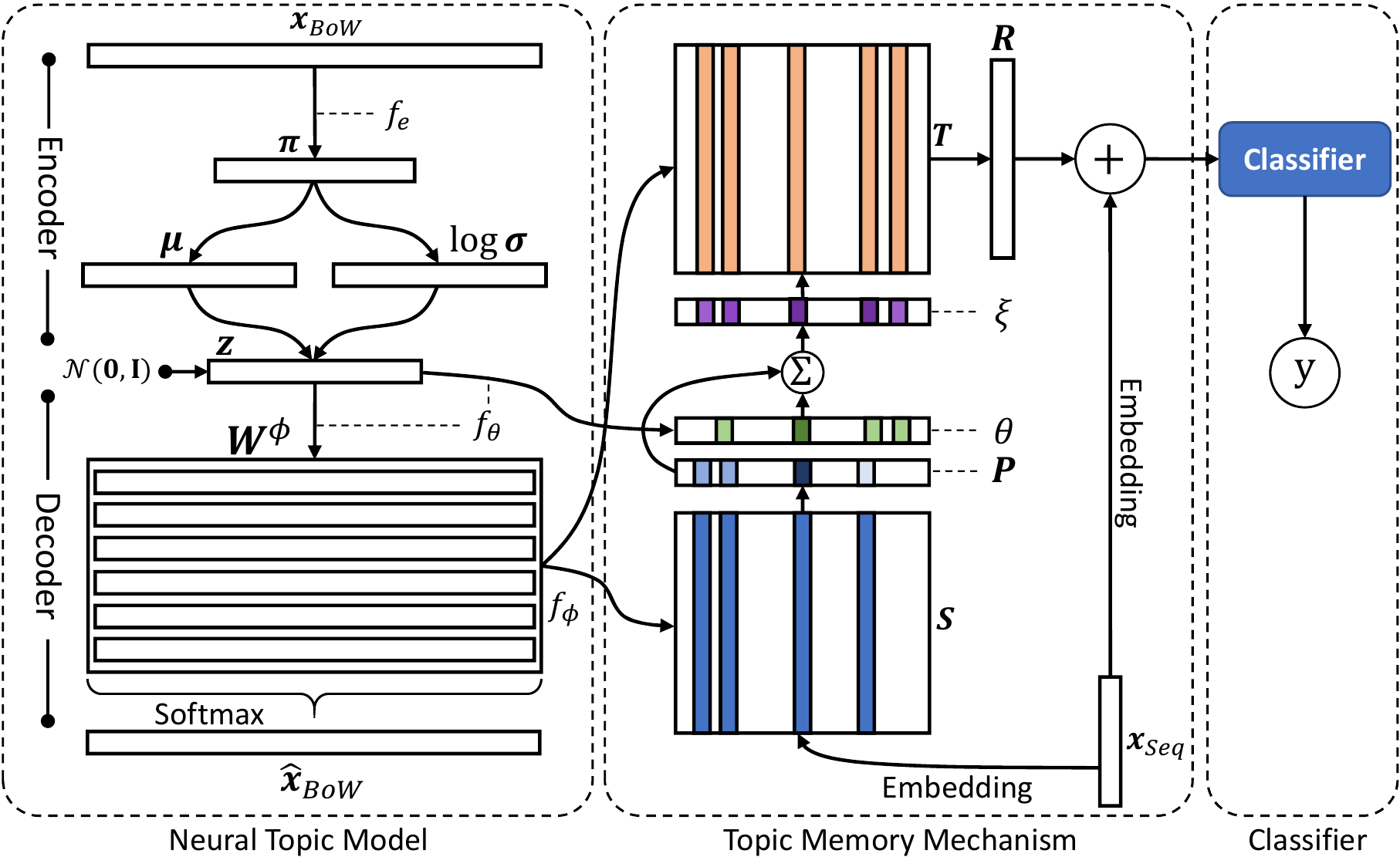}
	\caption{
    The overall framework of our topic memory networks. The dotted boxes from left to right show the neural topic model, the topic memory mechanism, and the classifier. Here the classifier allows multiple options and the details are left out.
    }
	\label{fig:framework}
    \vskip -0.5em
\end{figure}
\section{Topic Memory Networks}\label{sec:model}

In this section, we describe our topic memory networks (TMN), whose overall architecture is shown in Figure~\ref{fig:framework}. There are three major components: (1) a neural topic model (NTM) to induce latent topics (described in Section~\ref{ssec:model:ntm}), (2) a topic memory mechanism that maps the inferred latent topics to classification features (described in Section~\ref{ssec:model:tmn}), and (3) a text classifier, which produces the final classification labels for instances. These three components can be updated simultaneously via a joint learning process, which is introduced in Section~\ref{ssec:model:multi-task}. In particular, for the classifier, our TMN framework allows the combination of multiple options, e.g., CNN and RNN, which can be determined by the specific application scenario.





Formally, given ${\bf X}=\{{\bf x}^1, {\bf x}^2, \dots, {\bf x}^M\}$ as the input with $M$ short text instances, each instance ${\bf x}$ is processed into two representations: bag-of-words (BoW) term vector ${\bf x}_{BoW}\in \mathbb{R}^{V}$ and word index sequence vector ${\bf x}_{Seq}\in \mathbb{R}^{L}$, where $V$ is the vocabulary size and $L$ is the sequence length. ${\bf x}_{BoW}$ is fed into the neural topic model to induce latent topics.
Such topics are further matched with the embedded ${\bf x}_{Seq}$ to learn classification features in the topic memory mechanism. Then, the classifier concatenates the representations produced by the topic memory mechanism and the embedded ${\bf x}_{Seq}$ to predict the classification label $y$ for $\bf x$. 



\subsection{Neural Topic Model}\label{ssec:model:ntm}

Our topic model is inspired by neural topic model (NTM)~\cite{DBLP:conf/icml/MiaoGB17,srivastava2017autoencoding} that induces latent topics in neural networks. NTM is based on variational auto-encoder (VAE)~\cite{DBLP:journals/corr/KingmaW13}, involved with a continuous latent variable $\bf z$ as an intermediate representation. 
Here in NTM, the latent variable ${\bf z} \in \mathbb{R}^K$, where $K$ denotes the number of topics. 
In the following, we describe the generation and the inference of the model in turn.

\begin{figure}[t]
	\centering	\includegraphics[width=0.45 \textwidth]{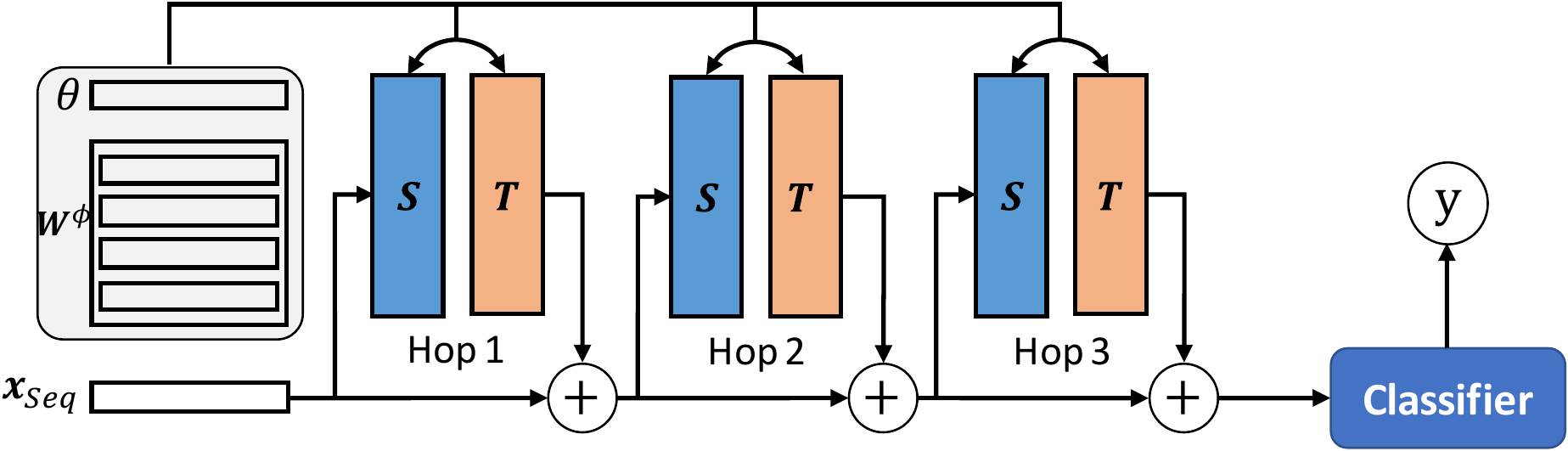}
    \vskip -0.5em
	\caption{Topic memory network with three hops.
    }	\label{fig:multihop}
    \vskip -0.5em
\end{figure}

\paragraph{NTM Generation.} 
Similar to LDA-style topic models, we assume $\bf x$ having a topic mixture $\theta$ represented as a $K$-dimensional distribution, which is generated via Gaussian softmax construction~\cite{DBLP:conf/icml/MiaoGB17}.
Each topic $k$ is represented by a word distribution $\phi_k$ over the vocabulary. 
Specifically, the generation story for $\bf x$ is:
\begin{compactitem}
\item Draw latent variable ${\bf z}\sim\mathcal{N}(\boldsymbol{\mu}, \boldsymbol{\sigma}^2)$ 
\item $\theta = \operatorname{softmax}(f_\theta ({\bf z}))$
\item For the $n$-th word in $\bf x$:
\begin{compactitem}
\item Draw word $w_{n}\sim \operatorname{softmax}(f_\phi ({\bf \theta}))$
\end{compactitem}
\end{compactitem}
%
where $f_*(\cdot)$ is a neural perceptron that linearly transforms inputs, activated by a non-linear transformation. Here we use rectified linear units (ReLUs)~\cite{DBLP:conf/icml/NairH10} as activate functions. The prior parameters of $\bf z$, $\boldsymbol{\mu}$ and $\boldsymbol{\sigma}$, are estimated from the input data and defined as:
\begin{equation}
\boldsymbol{\mu} = f_{\mu}(f_e({\bf x}_{BoW}))
,\, \log\boldsymbol{\sigma} = f_{\sigma}(f_e({\bf x}_{BoW}))
\end{equation}
%
Note that NTM is based on VAE, where an encoder estimates the prior parameters and a decoder describes the generation story. Compared with the basic VAE, NTM includes the additional 
distributional vectors $\theta$ and $\phi$, which can yield latent topic representations and thus ensuring their better interpretability in learning process~\cite{DBLP:conf/icml/MiaoGB17}.


\paragraph{NTM Inference.}
In NTM, we use variational inference~\cite{DBLP:journals/corr/BleiKM16} to approximate a posterior distribution over $\bf z$ given all the instances. The loss function of NTM is defined as
\begin{equation}
\mathcal{L}_{NTM} = D_{KL}(q({\bf z})\,||\,p({\bf z}\,|\,{\bf x})) - \mathbb{E}_{q({\bf z})}[p({\bf x}\,|\,{\bf z})]
\end{equation}
\noindent the negative of variational lower bound, where $q({\bf z})$ is a standard Normal prior $\mathcal{N}({\bf 0}, {\bf I})$. $p({\bf z}\,|\,{\bf x})$ and $p({\bf x}\,|\,{\bf z})$ are probabilities to describe encoding and decoding processes, respectively.\footnote{In implementation, to smooth the gradients, we apply reparameterization on $\bf z$ following previous work~\cite{DBLP:journals/corr/KingmaW13,DBLP:conf/icml/RezendeMW14}.}
Due to the space limitation, we leave out the derivation details and refer the readers to \citet{DBLP:conf/icml/MiaoGB17}.

\subsection{Topic Memory Mechanism}\label{ssec:model:tmn}
We exploit a topic memory mechanism to map the latent topics produced by NTM (described in Section~\ref{ssec:model:ntm}) to the features for classification. Inspired by memory networks \cite{DBLP:journals/corr/WestonCB14,DBLP:conf/nips/SukhbaatarSWF15}, we design two memory matrices, a source memory $\boldsymbol{S}$ and a target memory $\boldsymbol{T}$, both of which are in $K\times E$ size ($K$ for the number of topics and $E$ for the pre-defined size of word embeddings). $\boldsymbol{S}$ and $\boldsymbol{T}$ are produced by two ReLU-actived neural perceptrons, both taking the topic-word weight matrix $\boldsymbol{W}^{\phi}\in \mathbb{R}^{K\times V}$ as inputs. Recall that in NTM, we use $f_\phi(\cdot)$ to compute the word distributions given $\theta$. $\boldsymbol{W}^{\phi}$ is the kernel weight matrix of $f_\phi(\cdot)$, where $\boldsymbol{W}^{\phi}_{k,v}$ represents the importance of the $v$-th word in reflecting the $k$-th topic. Assuming $\boldsymbol{U}$ as the embedded $\mathbf{x}_{Seq}$ (word sequence form of $\bf x$), in source memory, we compute the match between the $k$-th topic and the embedding of the $l$-th word in $\mathbf{x}_{Seq}$ by
\begin{equation}\label{eq:mem}
\boldsymbol{P}_{k, l} = \operatorname{sigmoid}({\bf W}^{s}[\boldsymbol{S}_k;\boldsymbol{U}_l] + b^{s})
\end{equation}
\noindent where $[\bf x;\bf y]$ denotes the merge of $\bf x$ and $\bf y$, and we use concatenation operation here~\cite{DBLP:conf/emnlp/Dou17,DBLP:conf/emnlp/ChenSBY17}. ${\bf W}^{s}$ and $b^{s}$ are parameters to be learned. To further combine the instance-topic mixture $\theta$ with $\boldsymbol{P}$, we define the integrated memory weights as
\begin{equation}
\xi_k=\theta_k + \gamma\sum_l\boldsymbol{P}_{k,l}
\end{equation} 
\noindent where $\gamma$ is the pre-defined coefficient. Then, in target memory, via weighting target memory matrix $\boldsymbol{T}$ with $\xi$, we obtain the output representation $\boldsymbol{R}$ of the topic memory mechanism:
\begin{equation}
\boldsymbol{R}_k = \xi_k\boldsymbol{T}_k
\end{equation}
\noindent The concatenation of $\boldsymbol{R}$ and $\boldsymbol{U}$ (embedded $\mathbf{x}_{Seq}$)  further serves as the features for classification.

In particular, similar to the memory networks in prior research~\cite{DBLP:conf/nips/SukhbaatarSWF15,DBLP:conf/emnlp/ChenSBY17}, our model can be extended to handle multiple computation layers (hops). As shown in Figure~\ref{fig:multihop}, each hop contains a source matrix and a target matrix, and different hops are stacked following the way presented in~\citet{DBLP:conf/nips/SukhbaatarSWF15}.

\subsection{Joint Learning}\label{ssec:model:multi-task}

The entire TMN model integrates the three modules in Figure~\ref{fig:framework}, i.e., the neural topic model, the topic memory mechanism, and the classifier, which can be updated simultaneously in one framework. In doing so, we jointly tackle topic modeling and classification, and define the loss function of the overall framework to combine the two effects as following:
\begin{equation}
\mathcal{L} = \mathcal{L}_{NTM} + \lambda \mathcal{L}_{CLS}
\label{eq:o}
\end{equation}
%
\noindent where $\mathcal{L}_{NTM}$ represents the loss of NTM and $\mathcal{L}_{CLS}$ is the cross entropy reflecting classification loss. $\lambda$ is the trade-off parameter controlling the balance between topic model and classification.
\begin{table}\small
\centering
  \begin{tabular}{|l|rrrr|}
  \hline
  \multirow{2}{*}{\textbf{Dataset}}&\# of&\# of &Avg len& \multirow{2}{*}{Vocab size}\\
  &labels&docs&per doc&\\
  \hline 
  Snippets & 8 & 12,332 & 17 & 7,334 \\
  TagMyNews & 7 & 32,567 & 8 & 9,433 \\
  Twitter & 50 & 15,056 & 5 & 6,962 \\
  Weibo & 50 & 21,944 & 6 & 10,121 \\
  \hline
  \end{tabular}
\caption{Statistics of the experimental datasets. Labels refers to class labels. Avg len per doc refers to the average count of words in each document instance.}
\vskip -1.0em
\label{tab:data}
\end{table}

\begin{table*}\small
\centering
\label{exp_comp}
\begin{tabular}{|l|rr|rr|rr|rr|}
\hline
\multirow{2}{*}{\textbf{Models}} & \multicolumn{2}{c|}{Snippets} & \multicolumn{2}{c|}{TagMyNews} & \multicolumn{2}{c|}{Twitter} & \multicolumn{2}{c|}{Weibo} \\
\cline{2-9}
 & Acc & Avg F1 & Acc &  Avg F1 & Acc & Avg F1 & Acc & Avg F1 \\
\hline
\hline
\underline{\textbf{Comparison models}}&&&&&&&&\\
Majority Vote & 0.202 & 0.068 & 0.247 & 0.098 & 0.073 & 0.010 & 0.102 & 0.019 \\
SVM+BOW~\cite{DBLP:conf/acl/WangM12} & 0.210 & 0.080 & 0.259 & 0.058 & 0.070 & 0.009 & 0.116 & 0.039 \\
SVM+LDA~\cite{DBLP:journals/jmlr/BleiNJ03} & 0.689 & 0.694 & 0.616 & 0.593 & 0.159 & 0.111 & 0.192 & 0.147 \\
SVM+BTM~\cite{DBLP:conf/www/YanGLC13} & 0.772 & 0.772 & 0.686 & 0.677 & 0.232 & 0.164 & 0.331 & 0.277 \\
SVM+NTM~\cite{DBLP:conf/icml/MiaoGB17} & 0.779 & 0.776 & 0.664 & 0.654 & 0.261 & 0.177 & 0.379 & 0.348 \\
AttBiLSTM~\cite{DBLP:journals/corr/ZhangW15a} & 0.943 & 0.943 & 0.838 & 0.828 & 0.375 & 0.348 & 0.547 & 0.547 \\
CNN~\cite{DBLP:conf/emnlp/Kim14}		& 0.944 & 0.944 & 0.843 & 0.843 & 0.381 & 0.362 & 0.553 & 0.550 \\
CNN+TEWE~\cite{DBLP:conf/aaai/RenZZJ16a} 	& 0.944 & 0.944 & 0.846 & 0.846 & 0.385 & 0.368 & 0.537 & 0.532 \\
CNN+NTM  & 0.945 & 0.945 & 0.844 & 0.844 & 0.382 & 0.365 & 0.556 & 0.556 \\
\hline
\hline
\underline{\textbf{Our models}}&&&&&&&&\\
TMN (\textit{Separate TM Inference}) & 0.961 & 0.961 & 0.848 & 0.847 & 0.394 & \textbf{0.386} & 0.568 & 0.569\\
TMN (\textit{Joint TM Inference}) 	&  \textbf{0.964} & \textbf{0.964} &\textbf{0.851} & \textbf{0.851} & \textbf{0.397} & 0.375 & \textbf{0.591} & \textbf{0.589} \\
\hline           
\end{tabular}
\caption{Comparisons of accuracy (Acc) and average F1 (Avg F1) on four benchmark datasets. Our TMN, either with separate or joint TM inference, performs significantly better than all the
comparisons ($p < 0.05$, paired t-test).
}\label{tab:exp:classification}

\end{table*}




\section{Experiment Setup}


\subsection{Datasets}
We conduct experiments on four short text datasets, namely, Snippets, TagMyNews, Twitter, and Weibo.
Their details are described as follows.


\paragraph{Snippets.} This dataset contains Google search snippets released by~\citet{DBLP:conf/www/PhanNH08}. There are eight ground-truth labels, e.g., \textit{health} and \textit{sport}.


\paragraph{TagMyNews.}
 We use the news titles as instances from the benchmark classification dataset released by ~\citet{DBLP:conf/ecir/VitaleFS12}.\footnote{\url{http://acube.di.unipi.it/tmn-dataset/}} This dataset contains English news from really simple syndication (RSS) feeds. Each news feed (with its title) is annotated with one from seven labels, e.g., \textit{sci-tech}.

\paragraph{Twitter.} This dataset is 
used to evaluate tweet topic classification, which is 
built on the dataset released by TREC2011 microblog track.\footnote{\url{http://trec.nist.gov/data/tweets}}
Following previous settings~\cite{DBLP:conf/www/YanGLC13,DBLP:conf/sigir/LiWZSM16}, hashtags, i.e., user-annotated topic labels in each tweet such as ``\textit{\#Trump}'' and ``\textit{\#SuperBowl}'', serve as our ground-truth class labels.
Specifically, we construct the dataset with the following steps.
First, we remove the tweets without hashtags.
Second, we rank hashtags by their frequencies.
Third, we manually remove the hashtags that cannot mark topics, such as  ``\textit{\#fb}'' for indicating the source of tweets from Facebook, and combine the hashtags referring to the same topic, such as ``\textit{\#DonaldTrump}'' and ``\textit{\#Trump}''.
Finally, we select the top $50$ frequent hashtags, and all tweets containing these hashtags.




\paragraph{Weibo.} To evaluate our model on a different language other than English, we employ a Chinese dataset with short segments of text for topic classification.
This dataset is released by \citet{DBLP:conf/acl/LiLGHW16} with a collection of messages posted in June 2014 on Weibo, a popular Twitter alike platform in China.\footnote{The original dataset contains conversations to enrich the context of Weibo posts, which are not considered here.}
Similar to Twitter, Weibo allows up to $140$ Chinese characters in its messages.
In this Weibo dataset, each Weibo message is labeled with a hashtag as its category, and there are $50$ distinct hashtag labels in total, following the same procedure performed for the Twitter dataset.



Table~\ref{tab:data} shows the statistic information of the four datasets. Each dataset is randomly split into 80\% for training and 20\% for test.
20\% of randomly selected training instances are used to form development set.
We preprocess our English datasets, i.e., Snippets, TagMyNews, and Twitter, with \textsl{gensim tokenizer}\footnote{\url{https://radimrehurek.com/gensim/utils.html}} for tokenization. As to the Chinese Weibo dataset, we use FudanNLP toolkit~\cite{Qiu:2013}\footnote{\url{https://github.com/FudanNLP/fnlp}} for word segmentation. In addition, for each dataset, we maintain a vocabulary built based on the training set with removal of stop words\footnote{
\url{https://radimrehurek.com/gensim/parsing/preprocessing.html}} and words occurring less than $3$ times. The inputs of topic models ${\bf x}_{BoW}$ are constructed based on this vocabulary following common topic model settings~\cite{DBLP:journals/jmlr/BleiNJ03, DBLP:conf/icml/MiaoYB16}. Differently, we use the raw word sequence (without words removal) for the inputs of classification ${\bf x}_{Seq}$ as is done in previous work of text classification~\cite{DBLP:conf/emnlp/Kim14,DBLP:conf/acl/LiuQH17}.


\subsection{Model Settings}

We use pre-trained embeddings to initialize all word embeddings. For Snippets and TagMyNews datasets, we use pre-trained \texttt{GloVe} embeddings~\cite{DBLP:conf/emnlp/PenningtonSM14}\footnote{\url{http://nlp.stanford.edu/data/glove.6B.zip} (200d)}. For Twitter and Weibo datasets, we pre-train embeddings on large-scale external data with 99M tweets and 467M Weibo messages, respectively. 
For the number of topics, we follow previous settings~\cite{DBLP:conf/www/YanGLC13,DBLP:conf/acl/DasZD15,DBLP:journals/corr/DiengWGP16} to set $K=50$.
For all the other hyperparameters, we tune them on the development set by grid search. For our classifier, we employ CNN in experiment because of its better performance in short text classification than its counterparts such as RNN~\cite{DBLP:conf/ijcai/WangWZY17}.
The hidden size of CNN is set as $500$. The dimension of word embedding $E=200$. $\gamma=0.8$ for trading off $\theta$ and $\boldsymbol{P}$, and $\lambda=1.0$ for controlling the effects of topic model and classification. In the learning process, we run our model for $800$ epochs with early-stop strategy applied~\cite{DBLP:conf/nips/CaruanaLG00}.

\subsection{Comparison Models}
For comparison, we consider a weak baseline of majority vote, which assigns the major class labels in training set to all test instances. 
We further compare with the widely-used baseline SVM+BOW, SVM with unigram features~\cite{DBLP:conf/acl/WangM12}. We also consider other SVM-based baselines: SVM+LDA, SVM+BTM, SVM+NTM, whose features are topic distributions for instances learned by LDA~\cite{DBLP:journals/jmlr/BleiNJ03}, BTM~\cite{DBLP:conf/www/YanGLC13}, and NTM~\cite{DBLP:conf/icml/MiaoGB17}, respectively. In particular, BTM is one of the state-of-the-art topic models for short texts. 
To compare with neural classifiers, we test bidirectional long short-term memory with attention (AttBiLSTM) \cite{DBLP:conf/paclic/ZhangZHY15} and convolutional neural network (CNN) classifiers \cite{DBLP:conf/emnlp/Kim14}.
No topic representation is encoded in these two classifiers.
We also compare with the state-of-the-art short-text classifier CNN+TEWE~\cite{DBLP:conf/aaai/RenZZJ16a}, i.e., CNN classifier with topic-enriched word embeddings (TEWE), where the word embeddings are enriched by pre-trained NTM-inferred topic models. Moreover, to investigate the effectiveness of our proposed topic memory mechanism, we compare with CNN+NTM, which concatenates the representations learned by CNN and topics induced by NTM as classification features. In addition, we compare with our variant, TMN (\textit{Separate TM Inference}), where topics are induced separately before classification, and only used for initializing the topic memory. To be consistent, our model with a joint learning process for topic modeling and classification, described in Section \ref{ssec:model:multi-task}, is named as TMN (\textit{Joint TM Inference}). Note that the comparison CNN-based models share the same settings as our model, and the hidden size for each direction of BiLSTM is set to $100$.

\section{Experimental Results}

\subsection{Classification Comparison}\label{ssec:exp:classification}

Table~\ref{tab:exp:classification} shows the comparison on classification results, where the accuracy and average F1 scores on different classes labels are reported.
We have the following observations.

\paragraph{$\bullet$~\textit{Topic representations are indicative features.}} On all four datasets, simply by combining topic representations into features, SVM models produce better results than the models without exploiting topic features (\textit{i.e.}, SVM+BOW). 
This observation indicates that latent topic representations captured at corpus level are helpful to alleviate the data sparsity problem in short text classification.

\paragraph{$\bullet$~\textit{Neural network models are effective.}}
It is seen that neural models based on either CNN or AttBiLSTM yield better results than SVM.
This observation shows the effectiveness of representation learning in neural networks for short texts. 

\paragraph{$\bullet$~\textit{CNN serves as a better classifier for short texts than AttBiLSTM.}}
In comparison of CNN and AttBiLSTM without taking topic features, we observe that CNN yields generally better results on all the four datasets. This is consistent with the discovery in~\citet{DBLP:conf/ijcai/WangWZY17}, where CNN can better encode short texts than sequential models.

\paragraph{$\bullet$~\textit{Topic memory is useful to classification.}}
By exploring topic representations in memory mechanisms, our TMN model, inferring topic models either separately or jointly with classification, significantly outperform the best comparison models on each of the four datasets. 
Particularly, when compared with CNN+TEWE and CNN+NTM, both concatenating topics as part of the features, the results yielded by TMN are better. This demonstrates the effectiveness of topic memory to learn indicative topic representations for short text classification. 

\begin{table}\small
\centering
\begin{tabular}{|l|rrr|}
\hline
\textbf{Model}  & Snippets & TagMyNews & Twitter \\
\hline
LDA & 0.436 & 0.449    & 0.436   \\
BTM & 0.435 & 0.463    &  0.435   \\
NTM & 0.463 & 0.468    & 0.463   \\
TMN & \textbf{0.487}   & \textbf{0.499}	& \textbf{0.468}  \\
\hline
\end{tabular}
\caption{$C_V$ coherence scores for topics generated by various models. Higher is better. The best result in each column is in \textbf{bold}. }\label{tab:exp:cv}
\vskip -0.5em
\end{table}

\paragraph{$\bullet$~\textit{Jointly inferring latent topics is effective to text classification.}
}
In comparison between two TMN variants, TMN (\textit{Joint TM Inference}) produces better classification results, though large margin improvements are not observed
on the three English datasets, i.e., TagMyNews, Snippets, and Twitter.
This may be because the classifiers do not rely too much on high-quality latent topics,
since other features may be sufficient to indicate the labels, e.g., word positions in the instance.
As a result, better topic models, learned via jointly induced with classification, may not provide richer information for classification. Nevertheless, we notice that on Chinese Weibo dataset, the jointly trained topic model improves the accuracy and average F1 by $2.3$\% and $2.0$\%, respectively.
It may result from the prevalence of word order misuse in informal Weibo messages.
This mis-order phenomenon is common in Chinese and generally does not affect understanding.
The rich information conveyed by Chinese characters are capable of indicating semantic meanings of words even without correct orders~\cite{DBLP:journals/csl/QinCW16,DBLP:conf/emnlp/WangZZ17}.
As a result, the CNN classifier, which encodes orders of words, may also bring such mis-order noise to classification.
For these instances with mis-ordered words, a better topic model that learns text instances as unordered words, provides useful representations that compensate the loss of information in word orders and in turn improves the performance of text classification.

\begin{table}\small
\centering
\begin{tabular}{|l|m{6cm}|}
\hline
LDA & mubarak \textcolor{blue}{\uwave{bring}} \textcolor{blue}{\uwave{run}} obama democracy speech \textcolor{blue}{\uwave{believe}} regime power \textcolor{red}{\ul{bowl}} \\
\hline
BTM & mubarak egypt push internet people government \textcolor{blue}{\uwave{phone}}  hosni \textcolor{blue}{\uwave{need}} son \\
\hline
NTM & mubarak people egyptian egypt \textcolor{blue}{\uwave{stay}} \textcolor{blue}{\uwave{tomorrow}} protest news \textcolor{blue}{\uwave{phone}} protester   \\ 
\hline
TMN & mubarak protest protester tahrir square egyptian al jazeera repo cairo    \\ \hline
\end{tabular}
\caption{Top $10$ representative terms of the sample latent topics discovered by various topic models from Twitter dataset. We interpret the topics as ``\textit{Egyptian revolution of 2011}'' according to their word distributions. \textcolor{blue}{\uwave{Non-topic words}} are wave-underlined and in blue, and \textcolor{red}{\ul{off-topic words}} are underlined and in red.}\label{tab:topic-words}
\vskip -0.5em
\end{table}

\subsection{Topic Coherence Comparison}
In Section~\ref{ssec:exp:classification}, we find that TMN can significantly outperform comparison models on short text classification.
In this section, we study whether jointly learning topic models and classification can be helpful in producing coherent and meaningful topics. We use the $C_V$ metric~\cite{DBLP:conf/wsdm/RoderBH15} computed by Palmetto toolkit\footnote{\url{https://github.com/dice-group/Palmetto}} to evaluate the topic coherence, which has been shown to give the closest scores to human evaluation compared to other widely-used topic coherence metrics like NPMI~\cite{bouma2009normalized}. Table~\ref{tab:exp:cv} shows the comparison results of LDA, BTM, NTM, and TMN on the three English datasets.\footnote{In the rest of this paper, without otherwise indicated, TMN is used as a short form for TMN (\textit{Joint TM Inference}).} 
Note that we do not report $C_V$ scores for Chinese Weibo dataset as the Palmetto toolkit cannot process Chinese topics.

As can be seen, TMN yields higher $C_V$ scores by large margins than all others in comparison. This indicates that jointly exploring classification would be effective in producing coherent topics. The reason is that the supervision from classification labels can guide unsupervised topic models in discovering meaningful and interpretable topics. We also observe that NTM produces better results than LDA and BTM, which implies the effectiveness of inducing topic models by neural networks.

To further analyze the quality of yielded topics, Table~\ref{tab:topic-words} shows the top $10$ words of the sample latent topics reflecting ``\textit{Egyptian revolution of 2011}'' discovered by various models. We find that LDA yields off-topic word ``\textit{bowl}''. For the results of BTM and NTM, though we do not find off-topic words, non-topic words like ``\textit{need}'' and ``\textit{stay}'' are included.\footnote{Off-topic words are more likely to be interpreted to reflect other topics. Non-topic words cannot clearly indicate the corresponding topic.} The topic generated by TMN appears to be the best, which presents indicative words like ``\textit{tahrir}'' and ``\textit{cairo}'', for the event.
\begin{table}\small
\centering
\begin{tabular}{|l|>{\centering}m{25pt}>{\centering}m{37pt}>{\centering}m{25pt}p{25pt}|}
\hline
\# of Hops  & Snippets & TagMyNews & Twitter & Weibo \\ \hline
TMN-1H & 0.958    & 0.841     & 0.382   & 0.568 \\ 
TMN-2H & \textbf{0.964}    & 0.843     & 0.383   & 0.578 \\ 
TMN-3H & 0.962    & 0.845     & 0.384   & 0.581 \\ 
TMN-4H & 0.961    & 0.846     & 0.389   & 0.582 \\ 
TMN-5H & 0.960    & \textbf{0.851}     & \textbf{0.397}   & \textbf{0.591} \\ 
TMN-6H & 0.958    & 0.848     & 0.388   & 0.579 \\ \hline
\end{tabular}
\caption{The impact of the \# of hops on accuracy.}
\label{tab:exp:layer}
\end{table}
\subsection{Results with Varying Hyperparameters}

We further study the impact of two important hyperparameters in TMN, i.e., the hop number and the topic number, which will be discussed in turn.

\paragraph{Impact of Hop Numbers.}
Recall that Figure~\ref{fig:multihop} shows the capacity of TMN in combining multiple hops. Here we analyze the effects of hop numbers on the accuracy of TMN. Table~\ref{tab:exp:layer} reports the results, where $N$H refers to using $N$ hops ($N=1,2,...,6$). As can be seen, generally, TMN with $5$ hops achieves the best accuracy on most datasets except for Snippets dataset. We also observe that, although within a particular range, more hops can produce better accuracy, the increasing trends are not always monotonic. For example, TMN-6H always exhibits lower accuracy than TMN-5H. This observation implies that the overall representation ability of TMN is enhanced as the increasing complexity of the model via combining more hops.
However, this enhancement will reach saturation when the hop number exceeds a threshold, which is $5$ hops for most datasets in our experiment.

\begin{figure}[t]
	\centering	\includegraphics[width=0.48 \textwidth]{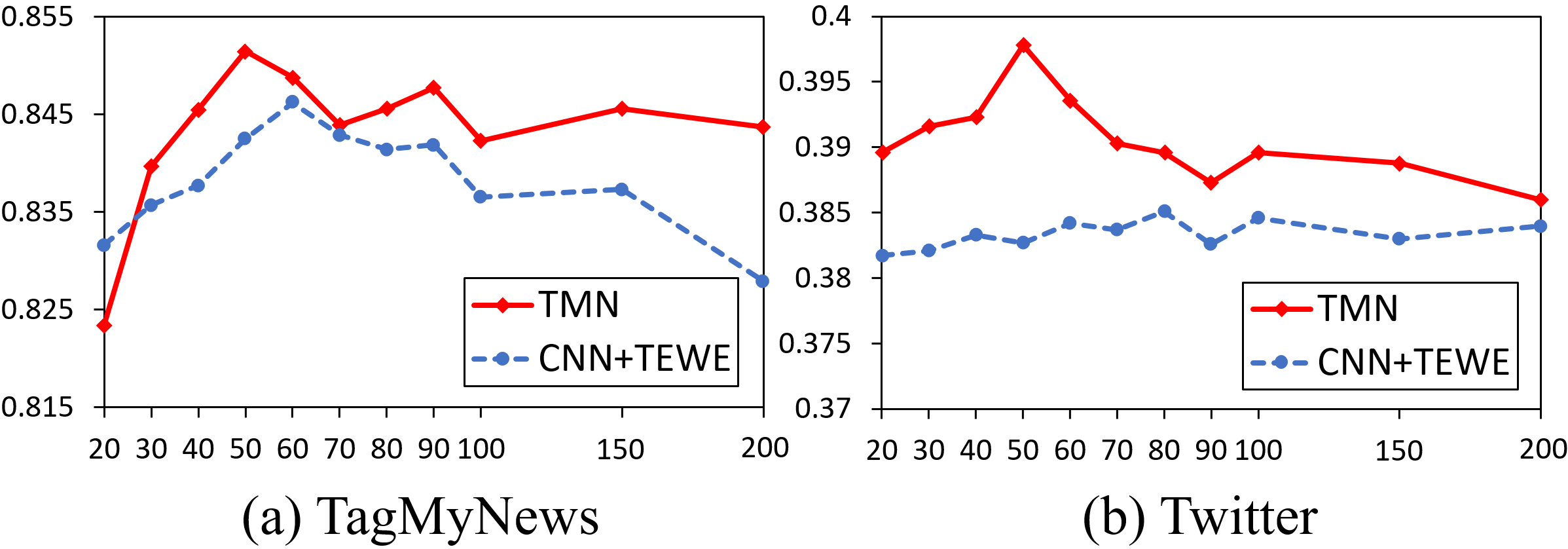}
    \caption{The impact of topic numbers, where the horizontal axis shows the number of topics and the vertical axis shows the accuracy.}
\label{fig:topic_num}
\end{figure}

\paragraph{Impact of Topic Numbers.} Figure \ref{fig:topic_num} shows the accuracy of TMN and CNN+TEWE (the best comparison model in Table \ref{tab:exp:classification}) given varying $K$, the number of topics on TagMyNews and Twitter datasets.\footnote{We observe similar distributions on Snippets and Weibo.} As we can see, the curves of all the models are not monotonic and the best accuracy is achieved given a particular number of topics, e.g., $K$=$50$ for TMN on TagMyNews dataset. When comparing different curves, we observe that TMN yields consistently better accuracy than CNN+TEWE, a comparison model shown in Table \ref{tab:exp:classification}, which demonstrates the robust performance of TMN over varying number of topics.


\begin{figure*}[ht]
	\centering
	\includegraphics[width=0.99 \textwidth, trim=0 20 0 0]{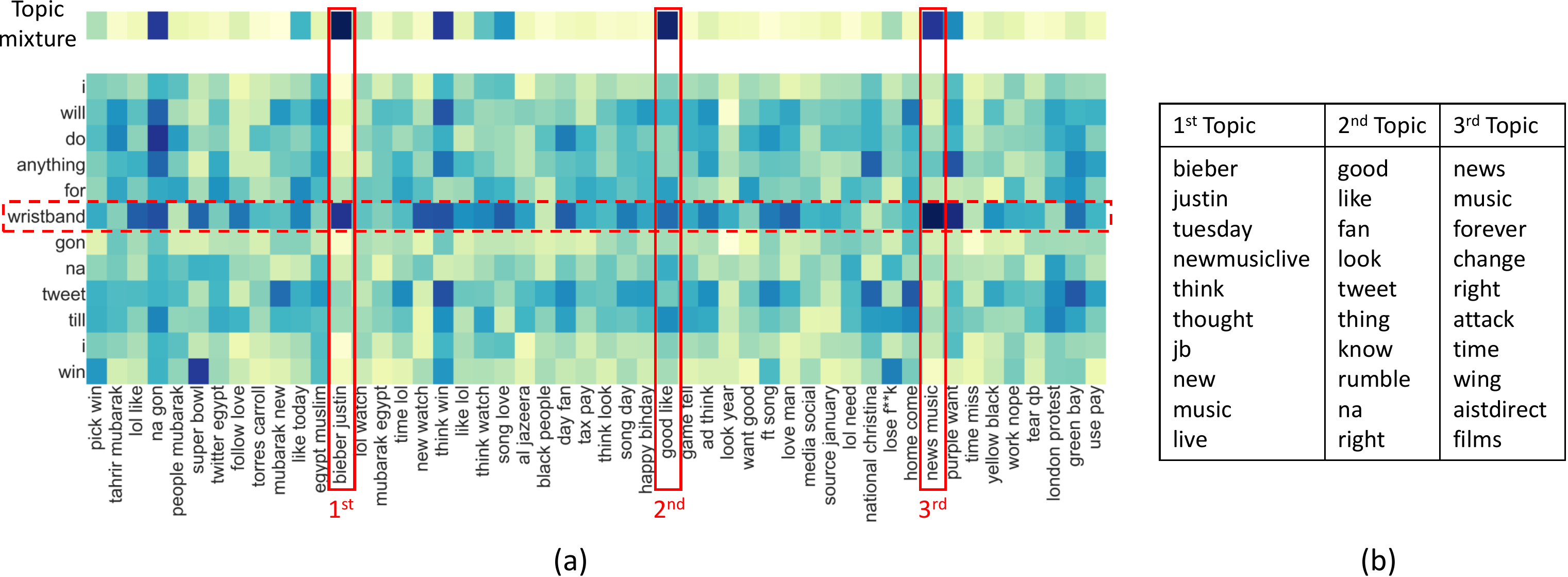}
	\vskip 0.5em
	\caption{Topic memory visualization for test instance S shown in Table~\ref{tab:intro-example}. (a) Heatmaps of topic mixture $\theta$ (the upper one) and topic memory weight matrix $\boldsymbol{P}$ (the lower one) illustrating the relevance between the words of S (left) and the learned topics (bottom, with top-2 words displayed). The red dotted rectangle indicates the representation for ``\textit{wristband}'', the topical word in S. The red rectangles with solid frames indicates the $3$ most relevant topics ordered by $\theta$. (b) Top-10 words of these topics indicated by $\phi$.}
\label{fig:mem_vis}
\end{figure*}
\subsection{A Case Study on Topic Memory}
Section~\ref{ssec:exp:classification} demonstrates the effectiveness of using topic memory on short text classification. To further understand why, in this section, we use the test instance S in Table~\ref{tab:intro-example} to analyze what the information captured by topic memory is indicative of class labels. Recall that the label of S, which should be \textsf{New.Music.Live}, can be indicated by containing word ``\textit{wristbands}'' and the collocation of ``\textit{wristbands}'' and 
``\textit{Bieber}'' in training instance R$_2$ labeled \textsf{New.Music.Live}. Figure~\ref{fig:mem_vis} shows the heatmaps of the weight matrix $\boldsymbol{P}$ in topic memory and the topic mixture $\theta$ captured by NTM for instance S. As can be seen, the top $3$ words for the latent topic with the largest value in $\theta$ are ``\textit{bieber}'', ``\textit{justine}'', and ``\textit{tuesday}'', which can effectively indicate the class label of S to be \textsl{New.Music.Live} because Justine Bieber was there on Tuesday. 
Interestingly, S contains none of the top three words. The latent semantic relations of S and these words are purely uncovered by the co-occurrence of words in S with other instances in the corpus, which further shows the benefit of using latent topics for alleviating the sparsity in short texts. We also observe that topic memory learns different representations for topical word ``\textit{wristband}'', highly indicating instance label, and background words, such as ``\textit{i}'' and ``\textit{for}''. This explains why topic memory is effective to classification.

\subsection{Error Analysis}
In this section, we take our classification results on TagMyNews dataset as an example to analyze our errors. We observe that one major type of incorrect prediction should be ascribed to the polysemy phenomenon. For example, the instance ``\textit{NBC gives `the voice' post super bowl slot}'' should be categorized as \textit{entertainment}. However, failing to understand the particular meaning of ``\textit{the voice}'' here as the name of a television singing competition, our model mistakenly categorizes this instance as \textsf{sport} because of the occurrence ``\textit{super bowl}''.
In future work, we would exploit context-sensitive topical word embeddings \cite{DBLP:conf/dexaw/WittSG16}, which is able to distinguish the meanings of the same word in different contexts.
Another main error type comes from the failure to capture phrase-level semantics. Taking ``\textit{On the merits of face time and living small}'' as an example, without understanding ``\textit{face time}'' as a phrase, our model wrongly predicts its category as \textsf{business} instead of its correct label as \textsf{sci\_tech}.
Such errors can be reduced by enhancing our NTM to phrase discovery topic models~\cite{DBLP:conf/emnlp/LindseyHS12,DBLP:conf/aaai/He16}, which is worthy exploring in future work.



\section{Related Work}\label{sec:related}

Our work mainly builds on two streams of prior work: short text classification and topic models.

\paragraph{Short Text Classification.} 
In the line of short text classification, most work focuses on alleviating the severe sparsity issues in short texts~\cite{DBLP:conf/www/YanGLC13}. Some previous efforts encode knowledge from external resource~\cite{DBLP:conf/cikm/JinLZYY11,DBLP:conf/cikm/LuciaF14,DBLP:conf/ijcai/WangWZY17,DBLP:conf/aaai/MaPC18}. Instead, our work 
learns effective representations only from internal data. For some specific classification tasks, such as sentiment analysis, manually-crafted features are designed to fit the target task~\cite{DBLP:conf/semeval/PakP10, DBLP:conf/acl/JiangYZLZ11}. Distinguished from them, we employ deep learning framework for representation learning, 
which requires no feature engineering process and thus ensures its general applicability to diverse classification scenarios. 
In comparison with the established classifiers applying deep learning methods
\cite{DBLP:conf/coling/SantosG14,DBLP:conf/naacl/LeeD16}, 
our work differs from them in the leverage of corpus-level latent topic representations for alleviating data sparsity issues.
In existing classification models using topic features, pre-trained topic mixtures are leveraged as part of features~\cite{DBLP:conf/www/PhanNH08,DBLP:conf/aaai/RenZZJ16a,DBLP:conf/emnlp/ChenSBY17}. Differently, our model encodes topic representations in a memory mechanism where topics are induced jointly with text classification in an end-to-end manner.

\paragraph{Topic Models.}
Well-known topic models, e.g., probabilistic latent
semantic analysis (pLSA)~\cite{DBLP:conf/sigir/Hofmann99} and latent Dirichlet allocation (LDA)~\cite{DBLP:journals/jmlr/BleiNJ03}, have shown advantages in capturing effective semantic representations, and proven beneficial to varying downstream applications, such as summarization~\cite{DBLP:conf/naacl/HaghighiV09} and recommendation~\cite{DBLP:conf/naacl/ZengLWBSW18,bai2018neural}.
For short text data, topic model variants have been proposed to reduce the effects of sparsity issues on topic modeling, such as biterm topic model (BTM)~\cite{DBLP:conf/www/YanGLC13} and LeadLDA~\cite{DBLP:conf/acl/LiLGHW16}. Recently, owing to the popularity of variational auto-encoder (VAE)~\cite{DBLP:journals/corr/KingmaW13}, it is able to induce latent topics in neural networks, namely, neural topic models (NTM)
\cite{DBLP:conf/icml/MiaoGB17,srivastava2017autoencoding}.
Although the concept of NTM has been mentioned earlier in~\citet{DBLP:conf/aaai/CaoLLLJ15}, their model is based on matrix factorization. Differently, VAE-style NTM~\cite{srivastava2017autoencoding,DBLP:conf/icml/MiaoGB17} follows the LDA fashion as 
probabilistic generative models, which is easy to interpret and extend. The NTM in our framework is in VAE-style, 
whose effects on short text classification serve as the key focus of our work.


\section{Conclusion}

We have presented topic memory networks that exploit corpus-level topic representations with a topic memory mechanism for short text classification.
The model alleviates data sparsity issues via jointly learning latent topics and text categories.
Empirical comparisons with state-of-the-art models on four benchmark datasets have demonstrated the validity and effectiveness of our model,
where better results have been achieved on both short text classification and topic coherence evaluation.

\section*{Acknowledgements}
This work is partially supported by the Research Grants Council of the Hong Kong Special Administrative Region, China (No. CUHK 14208815 and No. CUHK 14234416 of the General Research Fund), and Microsoft Research Asia (2018 Microsoft Research Asia Collaborative Research Award). We thank Shuming Shi, Dong Yu, Tong Zhang, Baolin Peng, Haoli Bai, and the three anonymous reviewers for the insightful suggestions on various aspects of this work.

\bibliographystyle{acl_natbib_nourl}
\bibliography{emnlp2018}

\begin{thebibliography}{55}
\expandafter\ifx\csname natexlab\endcsname\relax\def\natexlab#1{#1}\fi

\bibitem[{Bai et~al.(2018)Bai, Chen, Lyu, King, and Xu}]{bai2018neural}
Haoli Bai, Zhuangbin Chen, Michael~R. Lyu, Irwin King, and Zenglin Xu. 2018.
\newblock {Neural Relational Topic Model for Scientific Article Analysis}.
\newblock In \emph{Proceedings of the 27th ACM International Conference on
  Information and Knowledge Management, {CIKM} 2018}. ACM.

\bibitem[{Blei et~al.(2016)Blei, Kucukelbir, and
  McAuliffe}]{DBLP:journals/corr/BleiKM16}
David~M. Blei, Alp Kucukelbir, and Jon~D. McAuliffe. 2016.
\newblock {Variational Inference: {A} Review for Statisticians}.
\newblock \emph{CoRR}, abs/1601.00670.

\bibitem[{Blei et~al.(2003)Blei, Ng, and Jordan}]{DBLP:journals/jmlr/BleiNJ03}
David~M. Blei, Andrew~Y. Ng, and Michael~I. Jordan. 2003.
\newblock {Latent Dirichlet Allocation}.
\newblock \emph{Journal of Machine Learning Research}, 3:993--1022.

\bibitem[{Bouma(2009)}]{bouma2009normalized}
Gerlof Bouma. 2009.
\newblock {Normalized (Pointwise) Mutual Information in Collocation
  Extraction}.
\newblock \emph{Proceedings of GSCL}, pages 31--40.

\bibitem[{Cao et~al.(2015)Cao, Li, Liu, Li, and Ji}]{DBLP:conf/aaai/CaoLLLJ15}
Ziqiang Cao, Sujian Li, Yang Liu, Wenjie Li, and Heng Ji. 2015.
\newblock {A Novel Neural Topic Model and Its Supervised Extension}.
\newblock In \emph{Proceedings of the Twenty-Ninth {AAAI} Conference on
  Artificial Intelligence}, Austin, Texas, {USA}, pages 2210--2216.

\bibitem[{Caruana et~al.(2000)Caruana, Lawrence, and
  Giles}]{DBLP:conf/nips/CaruanaLG00}
Rich Caruana, Steve Lawrence, and C.~Lee Giles. 2000.
\newblock {Overfitting in Neural Nets: Backpropagation, Conjugate Gradient,
  an"d Early Stopping}.
\newblock In \emph{Advances in Neural Information Processing Systems 13, Papers
  from Neural Information Processing Systems, {NIPS} 2000}, Denver, CO, {USA},
  pages 402--408.

\bibitem[{Chen et~al.(2017)Chen, Sun, Bing, and
  Yang}]{DBLP:conf/emnlp/ChenSBY17}
Peng Chen, Zhongqian Sun, Lidong Bing, and Wei Yang. 2017.
\newblock {Recurrent Attention Network on Memory for Aspect Sentiment
  Analysis}.
\newblock In \emph{Proceedings of the 2017 Conference on Empirical Methods in
  Natural Language Processing, {EMNLP} 2017}, Copenhagen, Denmark, pages
  452--461.

\bibitem[{Das et~al.(2015)Das, Zaheer, and Dyer}]{DBLP:conf/acl/DasZD15}
Rajarshi Das, Manzil Zaheer, and Chris Dyer. 2015.
\newblock {Gaussian {LDA} for Topic Models with Word Embeddings}.
\newblock In \emph{Proceedings of the 53rd Annual Meeting of the Association
  for Computational Linguistics and the 7th International Joint Conference on
  Natural Language Processing of the Asian Federation of Natural Language
  Processing, {ACL} 2015, Volume 1: Long Papers}, Beijing, China, pages
  795--804.

\bibitem[{Dieng et~al.(2016)Dieng, Wang, Gao, and
  Paisley}]{DBLP:journals/corr/DiengWGP16}
Adji~B. Dieng, Chong Wang, Jianfeng Gao, and John~William Paisley. 2016.
\newblock {TopicRNN: {A} Recurrent Neural Network with Long-Range Semantic
  Dependency}.
\newblock \emph{CoRR}, abs/1611.01702.

\bibitem[{Dou(2017)}]{DBLP:conf/emnlp/Dou17}
Zi{-}Yi Dou. 2017.
\newblock {Capturing User and Product Information for Document Level Sentiment
  Analysis with Deep Memory Network}.
\newblock In \emph{Proceedings of the 2017 Conference on Empirical Methods in
  Natural Language Processing, {EMNLP} 2017}, Copenhagen, Denmark, pages
  521--526.

\bibitem[{Graves et~al.(2014)Graves, Wayne, and
  Danihelka}]{DBLP:journals/corr/GravesWD14}
Alex Graves, Greg Wayne, and Ivo Danihelka. 2014.
\newblock {Neural Turing Machines}.
\newblock \emph{CoRR}, abs/1410.5401.

\bibitem[{Haghighi and Vanderwende(2009)}]{DBLP:conf/naacl/HaghighiV09}
Aria Haghighi and Lucy Vanderwende. 2009.
\newblock Exploring content models for multi-document summarization.
\newblock In \emph{Proceedings of Human Language Technologies: The 2009 Annual
  Conference of the North American Chapter of the Association for Computational
  Linguistics, {HLT-NAACL} 2009}, pages 362--370, Boulder, Colorado, USA.

\bibitem[{He(2016)}]{DBLP:conf/aaai/He16}
Yulan He. 2016.
\newblock {Extracting Topical Phrases from Clinical Documents}.
\newblock In \emph{Proceedings of the Thirtieth {AAAI} Conference on Artificial
  Intelligence}, Phoenix, Arizona, {USA}, pages 2957--2963.

\bibitem[{Hofmann(1999)}]{DBLP:conf/sigir/Hofmann99}
Thomas Hofmann. 1999.
\newblock {Probabilistic Latent Semantic Indexing}.
\newblock In \emph{Proceedings of the 22nd Annual International {ACM} {SIGIR}
  Conference on Research and Development in Information Retrieval, {SIGIR}
  '99}, Berkeley, CA, {USA}, pages 50--57.

\bibitem[{Hu et~al.(2015)Hu, Chen, and Zhu}]{DBLP:conf/emnlp/HuCZ15}
Baotian Hu, Qingcai Chen, and Fangze Zhu. 2015.
\newblock {{LCSTS:} {A} Large Scale Chinese Short Text Summarization Dataset}.
\newblock In \emph{Proceedings of the 2015 Conference on Empirical Methods in
  Natural Language Processing, {EMNLP} 2015}, Lisbon, Portugal, pages
  1967--1972.

\bibitem[{Jiang et~al.(2011)Jiang, Yu, Zhou, Liu, and
  Zhao}]{DBLP:conf/acl/JiangYZLZ11}
Long Jiang, Mo~Yu, Ming Zhou, Xiaohua Liu, and Tiejun Zhao. 2011.
\newblock {Target-dependent Twitter Sentiment Classification}.
\newblock In \emph{The 49th Annual Meeting of the Association for Computational
  Linguistics: Human Language Technologies, Proceedings of the Conference},
  Portland, Oregon, {USA}, pages 151--160.

\bibitem[{Jin et~al.(2011)Jin, Liu, Zhao, Yu, and
  Yang}]{DBLP:conf/cikm/JinLZYY11}
Ou~Jin, Nathan~Nan Liu, Kai Zhao, Yong Yu, and Qiang Yang. 2011.
\newblock {Transferring Topical Knowledge from Auxiliary Long Texts for Short
  Text Clustering}.
\newblock In \emph{Proceedings of the 20th {ACM} Conference on Information and
  Knowledge Management, {CIKM} 2011}, Glasgow, United Kingdom, pages 775--784.

\bibitem[{Joulin et~al.(2017)Joulin, Grave, Bojanowski, and
  Mikolov}]{joulin2017bag}
Armand Joulin, Edouard Grave, Piotr Bojanowski, and Tomas Mikolov. 2017.
\newblock {Bag of Tricks for Efficient Text Classification}.
\newblock In \emph{Proceedings of the 15th Conference of the European Chapter
  of the Association for Computational Linguistics: Volume 2, Short Papers},
  pages 427--431.

\bibitem[{Kim(2014)}]{DBLP:conf/emnlp/Kim14}
Yoon Kim. 2014.
\newblock {Convolutional Neural Networks for Sentence Classification}.
\newblock In \emph{Proceedings of the 2014 Conference on Empirical Methods in
  Natural Language Processing, {EMNLP} 2014, {A} meeting of SIGDAT, a Special
  Interest Group of the {ACL}}, Doha, Qatar, pages 1746--1751.

\bibitem[{Kingma and Welling(2013)}]{DBLP:journals/corr/KingmaW13}
Diederik~P. Kingma and Max Welling. 2013.
\newblock {Auto-Encoding Variational Bayes}.
\newblock \emph{CoRR}, abs/1312.6114.

\bibitem[{Lee and Dernoncourt(2016)}]{DBLP:conf/naacl/LeeD16}
Ji~Young Lee and Franck Dernoncourt. 2016.
\newblock {Sequential Short-Text Classification with Recurrent and
  Convolutional Neural Networks}.
\newblock In \emph{The 2016 Conference of the North American Chapter of the
  Association for Computational Linguistics: Human Language Technologies,
  {NAACL} {HLT} 2016}, San Diego California, USA, pages 515--520.

\bibitem[{Li et~al.(2016{\natexlab{a}})Li, Wang, Zhang, Sun, and
  Ma}]{DBLP:conf/sigir/LiWZSM16}
Chenliang Li, Haoran Wang, Zhiqian Zhang, Aixin Sun, and Zongyang Ma.
  2016{\natexlab{a}}.
\newblock {Topic Modeling for Short Texts with Auxiliary Word Embeddings}.
\newblock In \emph{Proceedings of the 39th International {ACM} {SIGIR}
  conference on Research and Development in Information Retrieval, {SIGIR}
  2016}, Pisa, Italy, pages 165--174.

\bibitem[{Li et~al.(2016{\natexlab{b}})Li, Liao, Gao, He, and
  Wong}]{DBLP:conf/acl/LiLGHW16}
Jing Li, Ming Liao, Wei Gao, Yulan He, and Kam{-}Fai Wong. 2016{\natexlab{b}}.
\newblock {Topic Extraction from Microblog Posts Using Conversation
  Structures}.
\newblock In \emph{Proceedings of the 54th Annual Meeting of the Association
  for Computational Linguistics, {ACL} 2016, Volume 1: Long Papers}, Berlin,
  Germany.

\bibitem[{Lindsey et~al.(2012)Lindsey, Headden, and
  Stipicevic}]{DBLP:conf/emnlp/LindseyHS12}
Robert~V. Lindsey, William Headden, and Michael Stipicevic. 2012.
\newblock {A Phrase-Discovering Topic Model Using Hierarchical Pitman-Yor
  Processes}.
\newblock In \emph{Proceedings of the 2012 Joint Conference on Empirical
  Methods in Natural Language Processing and Computational Natural Language
  Learning, EMNLP-CoNLL 2012}, Jeju Island, Korea, pages 214--222.

\bibitem[{Liu et~al.(2017)Liu, Qiu, and Huang}]{DBLP:conf/acl/LiuQH17}
Pengfei Liu, Xipeng Qiu, and Xuanjing Huang. 2017.
\newblock {Adversarial Multi-task Learning for Text Classification}.
\newblock In \emph{Proceedings of the 55th Annual Meeting of the Association
  for Computational Linguistics, {ACL} 2017, Volume 1: Long Papers}, Vancouver,
  Canada, pages 1--10.

\bibitem[{Lucia and Ferrari(2014)}]{DBLP:conf/cikm/LuciaF14}
William Lucia and Elena Ferrari. 2014.
\newblock {EgoCentric: Ego Networks for Knowledge-based Short Text
  Classification}.
\newblock In \emph{Proceedings of the 23rd {ACM} International Conference on
  Conference on Information and Knowledge Management, {CIKM} 2014}, Shanghai,
  China, pages 1079--1088.

\bibitem[{Ma et~al.(2018)Ma, Peng, and Cambria}]{DBLP:conf/aaai/MaPC18}
Yukun Ma, Haiyun Peng, and Erik Cambria. 2018.
\newblock {Targeted Aspect-Based Sentiment Analysis via Embedding Commonsense
  Knowledge into an Attentive {LSTM}}.
\newblock In \emph{Proceedings of the Thirty-Second {AAAI} Conference on
  Artificial Intelligence}, New Orleans, Louisiana, USA.

\bibitem[{Miao et~al.(2017)Miao, Grefenstette, and
  Blunsom}]{DBLP:conf/icml/MiaoGB17}
Yishu Miao, Edward Grefenstette, and Phil Blunsom. 2017.
\newblock {Discovering Discrete Latent Topics with Neural Variational
  Inference}.
\newblock In \emph{Proceedings of the 34th International Conference on Machine
  Learning, {ICML} 2017}, Sydney, NSW, Australia, pages 2410--2419.

\bibitem[{Miao et~al.(2016)Miao, Yu, and Blunsom}]{DBLP:conf/icml/MiaoYB16}
Yishu Miao, Lei Yu, and Phil Blunsom. 2016.
\newblock {Neural Variational Inference for Text Processing}.
\newblock In \emph{Proceedings of the 33nd International Conference on Machine
  Learning, {ICML} 2016}, New York City, NY, USA, pages 1727--1736.

\bibitem[{Nair and Hinton(2010)}]{DBLP:conf/icml/NairH10}
Vinod Nair and Geoffrey~E. Hinton. 2010.
\newblock {Rectified Linear Units Improve Restricted Boltzmann Machines}.
\newblock In \emph{Proceedings of the 27th International Conference on Machine
  Learning (ICML-10)}, Haifa, Israel, pages 807--814.

\bibitem[{Pak and Paroubek(2010)}]{DBLP:conf/semeval/PakP10}
Alexander Pak and Patrick Paroubek. 2010.
\newblock {Twitter Based System: Using Twitter for Disambiguating Sentiment
  Ambiguous Adjectives}.
\newblock In \emph{Proceedings of the 5th International Workshop on Semantic
  Evaluation, SemEval@ACL 2010}, Uppsala University, Uppsala, Sweden, pages
  436--439.

\bibitem[{Pennington et~al.(2014)Pennington, Socher, and
  Manning}]{DBLP:conf/emnlp/PenningtonSM14}
Jeffrey Pennington, Richard Socher, and Christopher~D. Manning. 2014.
\newblock {Glove: Global Vectors for Word Representation}.
\newblock In \emph{Proceedings of the 2014 Conference on Empirical Methods in
  Natural Language Processing, {EMNLP} 2014, {A} meeting of SIGDAT, a Special
  Interest Group of the {ACL}}, Doha, Qatar, pages 1532--1543.

\bibitem[{Phan et~al.(2008)Phan, Nguyen, and
  Horiguchi}]{DBLP:conf/www/PhanNH08}
Xuan~Hieu Phan, Minh~Le Nguyen, and Susumu Horiguchi. 2008.
\newblock {Learning to Classify Short and Sparse Text {\&} Web with Hidden
  Topics from Large-scale Data Collections}.
\newblock In \emph{Proceedings of the 17th International Conference on World
  Wide Web, {WWW} 2008}, Beijing, China, pages 91--100.

\bibitem[{Qin et~al.(2016)Qin, Cong, and Wan}]{DBLP:journals/csl/QinCW16}
Zengchang Qin, Yonghui Cong, and Tao Wan. 2016.
\newblock {Topic Modeling of Chinese Language beyond a Bag-of-Words}.
\newblock \emph{Computer Speech {\&} Language}, 40:60--78.

\bibitem[{Qiu et~al.(2013)Qiu, Zhang, and Huang}]{Qiu:2013}
Xipeng Qiu, Qi~Zhang, and Xuanjing Huang. 2013.
\newblock {FudanNLP: A Toolkit for Chinese Natural Language Processing}.
\newblock In \emph{Proceedings of Annual Meeting of the Association for
  Computational Linguistics}.

\bibitem[{Ren et~al.(2016)Ren, Zhang, Zhang, and Ji}]{DBLP:conf/aaai/RenZZJ16a}
Yafeng Ren, Yue Zhang, Meishan Zhang, and Donghong Ji. 2016.
\newblock {Improving Twitter Sentiment Classification Using Topic-Enriched
  Multi-Prototype Word Embeddings}.
\newblock In \emph{Proceedings of the Thirtieth {AAAI} Conference on Artificial
  Intelligence}, Phoenix, Arizona, {USA}, pages 3038--3044.

\bibitem[{Rezende et~al.(2014)Rezende, Mohamed, and
  Wierstra}]{DBLP:conf/icml/RezendeMW14}
Danilo~Jimenez Rezende, Shakir Mohamed, and Daan Wierstra. 2014.
\newblock {Stochastic Backpropagation and Approximate Inference in Deep
  Generative Models}.
\newblock In \emph{Proceedings of the 31th International Conference on Machine
  Learning, {ICML} 2014}, Beijing, China, pages 1278--1286.

\bibitem[{R{\"{o}}der et~al.(2015)R{\"{o}}der, Both, and
  Hinneburg}]{DBLP:conf/wsdm/RoderBH15}
Michael R{\"{o}}der, Andreas Both, and Alexander Hinneburg. 2015.
\newblock {Exploring the Space of Topic Coherence Measures}.
\newblock In \emph{Proceedings of the Eighth {ACM} International Conference on
  Web Search and Data Mining, {WSDM} 2015}, Shanghai, China, pages 399--408.

\bibitem[{dos Santos and Gatti(2014)}]{DBLP:conf/coling/SantosG14}
C{\'{\i}}cero~Nogueira dos Santos and Maira Gatti. 2014.
\newblock {Deep Convolutional Neural Networks for Sentiment Analysis of Short
  Texts}.
\newblock In \emph{25th International Conference on Computational Linguistics,
  Proceedings of the Conference: Technical Papers, {COLING} 2014}, Dublin,
  Ireland, pages 69--78.

\bibitem[{Srivastava and Sutton(2017)}]{srivastava2017autoencoding}
Akash Srivastava and Charles Sutton. 2017.
\newblock {Autoencoding Variational Inference For Topic Models}.
\newblock \emph{arXiv preprint arXiv:1703.01488}.

\bibitem[{Sukhbaatar et~al.(2015)Sukhbaatar, Szlam, Weston, and
  Fergus}]{DBLP:conf/nips/SukhbaatarSWF15}
Sainbayar Sukhbaatar, Arthur Szlam, Jason Weston, and Rob Fergus. 2015.
\newblock {End-To-End Memory Networks}.
\newblock In \emph{Advances in Neural Information Processing Systems 28: Annual
  Conference on Neural Information Processing Systems, {NIPS} 2015}, Montreal,
  Quebec, Canada, pages 2440--2448.

\bibitem[{Vitale et~al.(2012)Vitale, Ferragina, and
  Scaiella}]{DBLP:conf/ecir/VitaleFS12}
Daniele Vitale, Paolo Ferragina, and Ugo Scaiella. 2012.
\newblock {Classification of Short Texts by Deploying Topical Annotations}.
\newblock In \emph{Advances in Information Retrieval - 34th European Conference
  on {IR} Research, {ECIR} 2012}, Barcelona, Spain, pages 376--387.

\bibitem[{Wang et~al.(2017{\natexlab{a}})Wang, Wang, Zhang, and
  Yan}]{DBLP:conf/ijcai/WangWZY17}
Jin Wang, Zhongyuan Wang, Dawei Zhang, and Jun Yan. 2017{\natexlab{a}}.
\newblock {Combining Knowledge with Deep Convolutional Neural Networks for
  Short Text Classification}.
\newblock In \emph{Proceedings of the Twenty-Sixth International Joint
  Conference on Artificial Intelligence, {IJCAI} 2017}, Melbourne, Australia,
  pages 2915--2921.

\bibitem[{Wang et~al.(2017{\natexlab{b}})Wang, Zhang, and
  Zong}]{DBLP:conf/emnlp/WangZZ17}
Shaonan Wang, Jiajun Zhang, and Chengqing Zong. 2017{\natexlab{b}}.
\newblock {Exploiting Word Internal Structures for Generic Chinese Sentence
  Representation}.
\newblock In \emph{Proceedings of the 2017 Conference on Empirical Methods in
  Natural Language Processing, {EMNLP} 2017}, Copenhagen, Denmark, pages
  298--303.

\bibitem[{Wang and Manning(2012)}]{DBLP:conf/acl/WangM12}
Sida~I. Wang and Christopher~D. Manning. 2012.
\newblock {Baselines and Bigrams: Simple, Good Sentiment and Topic
  Classification}.
\newblock In \emph{The 50th Annual Meeting of the Association for Computational
  Linguistics, Proceedings of the Conference, {ACL} 2012, Volume 2: Short
  Papers}, Jeju Island, Korea, pages 90--94.

\bibitem[{Weston et~al.(2014)Weston, Chopra, and
  Bordes}]{DBLP:journals/corr/WestonCB14}
Jason Weston, Sumit Chopra, and Antoine Bordes. 2014.
\newblock {Memory Networks}.
\newblock \emph{CoRR}, abs/1410.3916.

\bibitem[{Witt et~al.(2016)Witt, Seifert, and
  Granitzer}]{DBLP:conf/dexaw/WittSG16}
Nils Witt, Christin Seifert, and Michael Granitzer. 2016.
\newblock {Explaining Topical Distances Using Word Embeddings}.
\newblock In \emph{27th International Workshop on Database and Expert Systems
  Applications, {DEXA} 2016 Workshops}, Porto, Portugal, pages 212--217.

\bibitem[{Xiao and Cho(2016)}]{DBLP:journals/corr/XiaoC16}
Yijun Xiao and Kyunghyun Cho. 2016.
\newblock {Efficient Character-level Document Classification by Combining
  Convolution and Recurrent Layers}.
\newblock \emph{CoRR}, abs/1602.00367.

\bibitem[{Yan et~al.(2013)Yan, Guo, Lan, and Cheng}]{DBLP:conf/www/YanGLC13}
Xiaohui Yan, Jiafeng Guo, Yanyan Lan, and Xueqi Cheng. 2013.
\newblock {A Biterm Topic Model for Short Sexts}.
\newblock In \emph{22nd International World Wide Web Conference, {WWW} 2013},
  Rio de Janeiro, Brazil, pages 1445--1456.

\bibitem[{Zeng et~al.(2018)Zeng, Li, Wang, Beauchamp, Shugars, and
  Wong}]{DBLP:conf/naacl/ZengLWBSW18}
Xingshan Zeng, Jing Li, Lu~Wang, Nicholas Beauchamp, Sarah Shugars, and
  Kam{-}Fai Wong. 2018.
\newblock Microblog conversation recommendation via joint modeling of topics
  and discourse.
\newblock In \emph{Proceedings of the 2018 Conference of the North American
  Chapter of the Association for Computational Linguistics: Human Language
  Technologies, {NAACL-HLT} 2018}, New Orleans, Louisiana, USA, pages 375--385.

\bibitem[{Zhang and Wang(2015)}]{DBLP:journals/corr/ZhangW15a}
Dongxu Zhang and Dong Wang. 2015.
\newblock {Relation Classification via Recurrent Neural Network}.
\newblock \emph{CoRR}, abs/1508.01006.

\bibitem[{Zhang et~al.(2015{\natexlab{a}})Zhang, Zheng, Hu, and
  Yang}]{DBLP:conf/paclic/ZhangZHY15}
Shu Zhang, Dequan Zheng, Xinchen Hu, and Ming Yang. 2015{\natexlab{a}}.
\newblock {Bidirectional Long Short-Term Memory Networks for Relation
  Classification}.
\newblock In \emph{Proceedings of the 29th Pacific Asia Conference on Language,
  Information and Computation, {PACLIC} 29}, Shanghai, China.

\bibitem[{Zhang et~al.(2012)Zhang, Wang, Xue, and
  Zha}]{DBLP:journals/tist/ZhangWXZ12}
Weinan Zhang, Dingquan Wang, Gui{-}Rong Xue, and Hongyuan Zha. 2012.
\newblock {Advertising Keywords Recommendation for Short-Text Web Pages Using
  Wikipedia}.
\newblock \emph{{ACM} {TIST}}, 3(2):36:1--36:25.

\bibitem[{Zhang et~al.(2015{\natexlab{b}})Zhang, Zhao, and
  LeCun}]{DBLP:conf/nips/ZhangZL15}
Xiang Zhang, Junbo~Jake Zhao, and Yann LeCun. 2015{\natexlab{b}}.
\newblock {Character-level Convolutional Networks for Text Classification}.
\newblock In \emph{Advances in Neural Information Processing Systems 28: Annual
  Conference on Neural Information Processing Systems, {NIPS} 2015}, Montreal,
  Quebec, Canada, pages 649--657.

\bibitem[{Zhang et~al.(2018)Zhang, Li, Song, and
  Zhang}]{DBLP:conf/naacl/ZhangLSZ18}
Yingyi Zhang, Jing Li, Yan Song, and Chengzhi Zhang. 2018.
\newblock Encoding conversation context for neural keyphrase extraction from
  microblog posts.
\newblock In \emph{Proceedings of the 2018 Conference of the North American
  Chapter of the Association for Computational Linguistics: Human Language
  Technologies, {NAACL-HLT} 2018}, New Orleans, Louisiana, USA, pages
  1676--1686.

\end{thebibliography}

\end{document}